\documentclass{svproc}
\usepackage{graphicx}
\usepackage{marvosym}
\usepackage{listings}
\usepackage{algorithm2e}
\RestyleAlgo{ruled}
\usepackage{comment}
\lstdefinelanguage{PDDL}
{
  sensitive=false,    
  morecomment=[l]{;}, 
  alsoletter={:,-},   
  morekeywords={
    define,domain,problem,not,and,or,when,forall,exists,either,
    :domain,:requirements,:types,:objects,:constants,
    :predicates,:action,:parameters,:precondition,:effect,
    :fluents,:primary-effect,:side-effect,:init,:goal,
    :strips,:adl,:equality,:typing,:conditional-effects,
    :negative-preconditions,:disjunctive-preconditions,
    :existential-preconditions,:universal-preconditions,:quantified-preconditions,
    :functions,assign,increase,decrease,scale-up,scale-down,
    :metric,minimize,maximize,
    :durative-actions,:duration-inequalities,:continuous-effects,
    :durative-action,:duration,:condition
  }
}
\usepackage{url}

\begin{document}
\mainmatter              
\title{Autonomous Electric Vehicle Battery Disassembly Based on NeuroSymbolic Computing}
\titlerunning{NeuroSymbolic Battery Disassembly}  
%
\author{Hengwei Zhang\inst{1,*} \and Hua Yang\inst{2,*} \and Haitao Wang\inst{2}  \and Zhigang Wang\inst{2} \and \\
Shengmin Zhang\inst{1} \and  Ming Chen\inst{1}\textsuperscript{(\Letter)}}
\authorrunning{Hengwei Zhang et al.} 
%
\tocauthor{Hengwei Zhang, Hua Yang, Haitao Wang, Zhigang Wang, Shengmin Zhang, Ming Chen}
\institute{Shanghai Jiaotong University Machinery \& Dynamic Engineering College, Shanghai, China,\\
\email{zhw\_SHJD827@sjtu.edu.cn, zhangshengmin@sjtu.edu.cn,\\ mingchen@sjtu.edu.cn},
\and
Intel Labs China,\\
\email{hua.yang@intel.com, hai.tao.wang@intel.com,\\ zhi.gang.wang@intel.com}\\
$\ast$ These authors contributed equally to this work}

\maketitle              

\begin{abstract}
The booming of electric vehicles demands efficient battery disassembly for recycling to be environment-friendly. Due to the unstructured environment and high uncertainties, battery disassembly is still primarily done by humans, probably assisted by robots. It is highly desirable to design autonomous solutions to improve work efficiency and lower human risks in high voltage and toxic environments. This paper proposes a novel framework of the NeuroSymbolic task and motion planning method to disassemble batteries in an unstructured environment using robots automatically. It enables robots to independently locate and disassemble battery bolts, with or without obstacles. This study not only provides a solution for intelligently disassembling electric vehicle batteries but also verifies its feasibility through a set of test results with the robot accomplishing the disassembly tasks in a complex and dynamic environment.
\end{abstract}
\section{Introduction}
With the booming of large-capacity lithium batteries, more automakers have adopted batteries to provide power to electric cars\cite{ref_1}. According to data from China Automotive Technology and Research Center (CATARC), in 2025, the amount of expired electric car batteries will reach 780k tons (about 116 GWh). End-of-life batteries pose hazards to the environment if not properly recycled. Leading-edge battery disassembly techniques rely on human-assisted machines, which not only are of low efficiency but also expose workers to hazardous working conditions. Having robots to disassemble batteries automatically is urgently needed to exempt humans from toxic working environments, improve cost efficiency, and handle heavy workloads that are rapidly growing.

As of now, major difficulties of using robots to autonomously disassemble batteries concern the unstructured working environment and a multitude of uncertainties. Automating the process of disassembly, recycling and sorting battery components cannot be done by pre-programming against a homogeneous batch since End-of-life batteries are often of different models and shapes \cite{ref_2}. For automatic disassembly, several researchers have been using computer vision-driven robots to carry out automatic disassembly tasks \cite{ref_3,ref_4}. However, these solutions assume good enough visual quality for robots to discern bolts, ignoring disassembly task planning.

\begin{figure}[h]
\centering
\includegraphics[width=12cm]{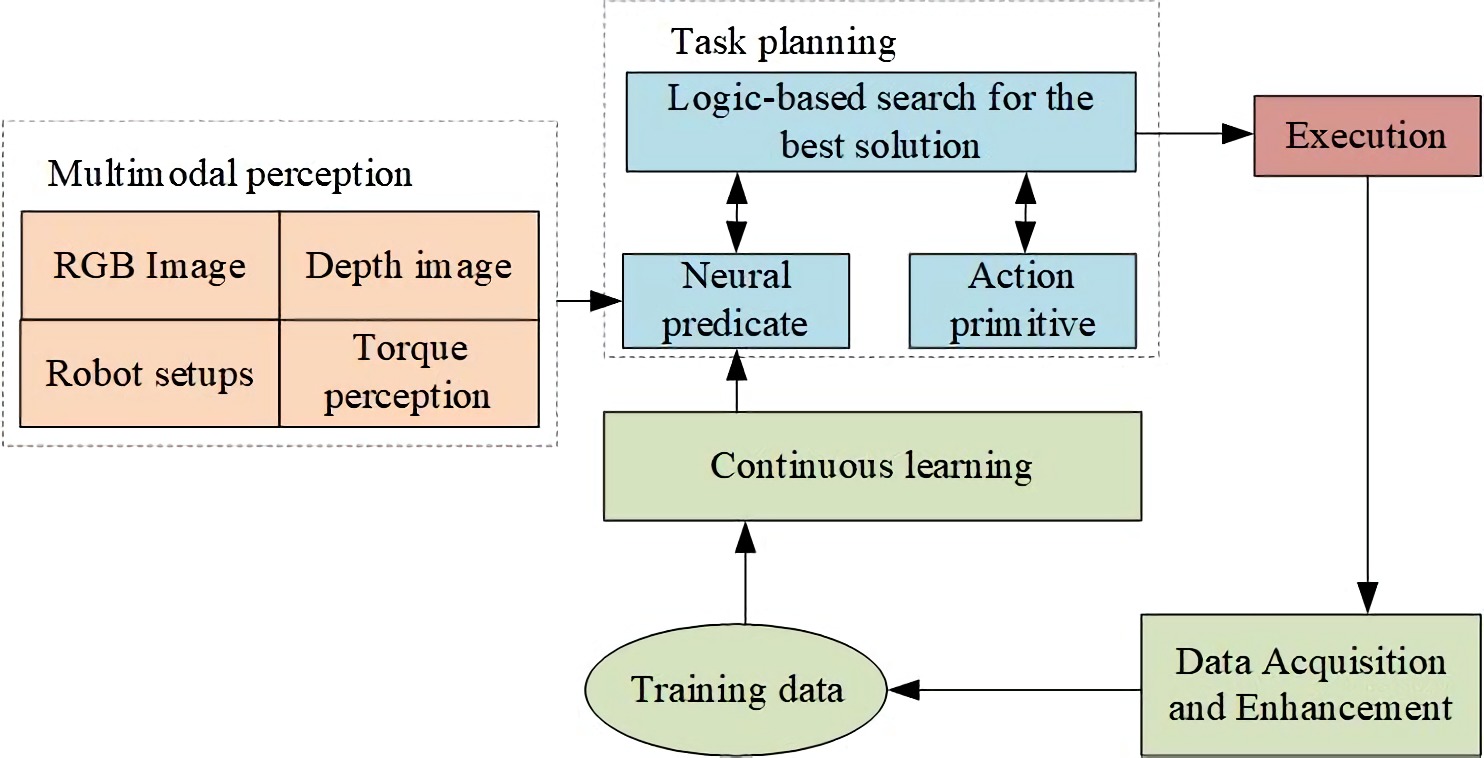}
\caption{Intelligent Disassembly system based on NeuroSymbolic.}
\label{fig:one}
\end{figure}

Figure \ref{fig:one} shows the architecture of the proposed NeuroSymbolic task and motion planning (NeuroSymbolic TAMP for short). It uses PDDL (planning domain define language) \cite{ref_5} to describe each disassembly primitive, which includes each action's preconditions and effect before and after execution, respectively. The optimum solution searching models rely on logical inference to automatically search for action primitive to reach the target state based on the current and target state. Unlike the classic planning method, we introduced neural predicates that use neural networks to ground symbols(convert continuous state, such as robots’ posture or visual inputs, to symbols required by planners). This method goes beyond logic symbols without requiring manual abstracting. Due to errors in the neural predicate output, a simple threshold method can easily lead to invalid states. We expand the search module and adopt probabilistic logic-based searching strategies \cite{ref_6,ref_7} to connect to the neural network’s softmax results.

This paper proposes a model that integrates inference capabilities from the symbolic logic system (AI 1.0) with the perception and learning capabilities from the neural network system (AI 2.0) \cite{ref_8} to carry out the disassembly task autonomously. The system can intelligently detect environment variation and autonomously choose and execute primitives for electrical electric vehicle(EV) battery disassembly tasks. It can be easily extended into more complex forms with new primitives. This paper implements a method of autonomously disassembling batteries in an unstructured and dynamic environment and proves its viability.

In this paper, we first summarize the current challenges during the battery disassembly process and briefly introduce the NeuroSymbolic TAMP in section 2. In section 3, we take bolt disassembly as a preliminary study and divide the disassembly tasks into five stages. The formulation of the disassembly task composed of primitives is proposed for subsequent content. Section 4 introduces the neural predicates and disassembly primitives, demonstrating the architecture and components of NeuroSymbolic TAMP in detail. Finally, we conduct a series of simulation tests to verify the effectiveness of the NeuroSymbolic TAMP. 

\section{State of the Art}
\subsection{Battery Disassembly}
Disassembly strategy study is one of the earliest researches for battery disassembly tasks, which currently are primarily carried out by humans \cite{ref_2,ref_3,ref_4}. From 2014 to 2015, researchers designed a disassembly workstation and conducted in-depth research on the Audi Q5 battery pack \cite{ref_17}. Recent research work is to further refine the problem, mainly according to the geometric structure of power battery, using genetic algorithm, matrix or graph theory to make decision and optimization on disassembly behavior, disassembly sequence and even disassembly trajectory \cite{ref_18,ref_19}. The research in this field has been mature, which provides scientific analysis and guidance for the efficient disassembly of power batteries. However, the specific implementation links still need to be completed manually.
Some researchers tried to use pre-programmed robots to automatically accomplish the task \cite{ref_20,ref_21}, which cannot be deployed extensively due to the highly dynamic disassembly environment and battery pack variations.

In addition to the above work, a lot of research has been done on robot task planning. Robot task planning refers to selecting a proper sequence of actions and constraints according to the initial state and target state of the manipulated object without interference, changing the state of the robot and the manipulated object, and realizing the transition from the initial state to the target state \cite{ref_22,ref_23}. 
Most of these methods carry out task planning based on artificial intelligence. First, they define different tasks and expand PDDL \cite{ref_5,ref_24} to achieve different solutions. For instance, in multi-stage complex tasks, the hierarchical planning method is adopted to decompose the overall robot task into simple tasks \cite{ref_25}. Furthermore, some scholars focus on motion planning and decision-making under uncertainty \cite{ref_26}, task planning and decision-making of robot pushing away from object under chaotic environment \cite{ref_27}, real-time online task planning and decision-making of intelligent robot operating object behavior \cite{ref_28} and so on. Although such research results are remarkable, they all have a common problem: the planner requires inputting the symbolic representation of the system state, which often needs to be pre-defined by semiotics experts, which significantly limits the application scope of task planning. At present, such research is still in the laboratory stage. 

Along with the development of deep learning, some researchers have tried to use computer vision neural networks to identify target types and locations, and then do motion planning for robots \cite{ref_4}. In addition, for task planning these methods usually assume accurate perception results which are either expensive to obtain or impractical due to stringent geometric accuracy requirements. Above all, the major obstacle towards disassembly is the inability of robots to plan based on perception results under an unstructured environment autonomously.  
\subsection{NeuroSymbolic Computing and Battery Disassembly}
NeuroSymbolic computing is a new technique that has the potential of organically integrating high-level inference with lower-level perception. The framework is of two levels: The high-level conducts knowledge-based logic inference, while the low level carries out probabilistic learning-based perception and control, etc. \cite{ref_9}. On one hand, it is imperative to design explicable and reliable task planning based on real-time information, which demands symbolic inference capabilities. On the other hand, it is impractical to symbolize all disassembly status \cite{ref_10}, which is also due to complicated battery disassembly scenarios. This has made probabilistic learning methods indispensable to learning symbolic representation from visual inputs. The high-level symbolic inference system can narrow down contexts for low-level perception, which greatly reduces the perception model complexities and expands its applicability in practical industrial scenarios \cite{ref_11,ref_12}. Based on these understandings, we use NeuroSymbolic methods to shed light on autonomous battery disassembly solutions.
\section{Problem Definition}
\begin{figure}[h]
\centering
\includegraphics[width=12cm]{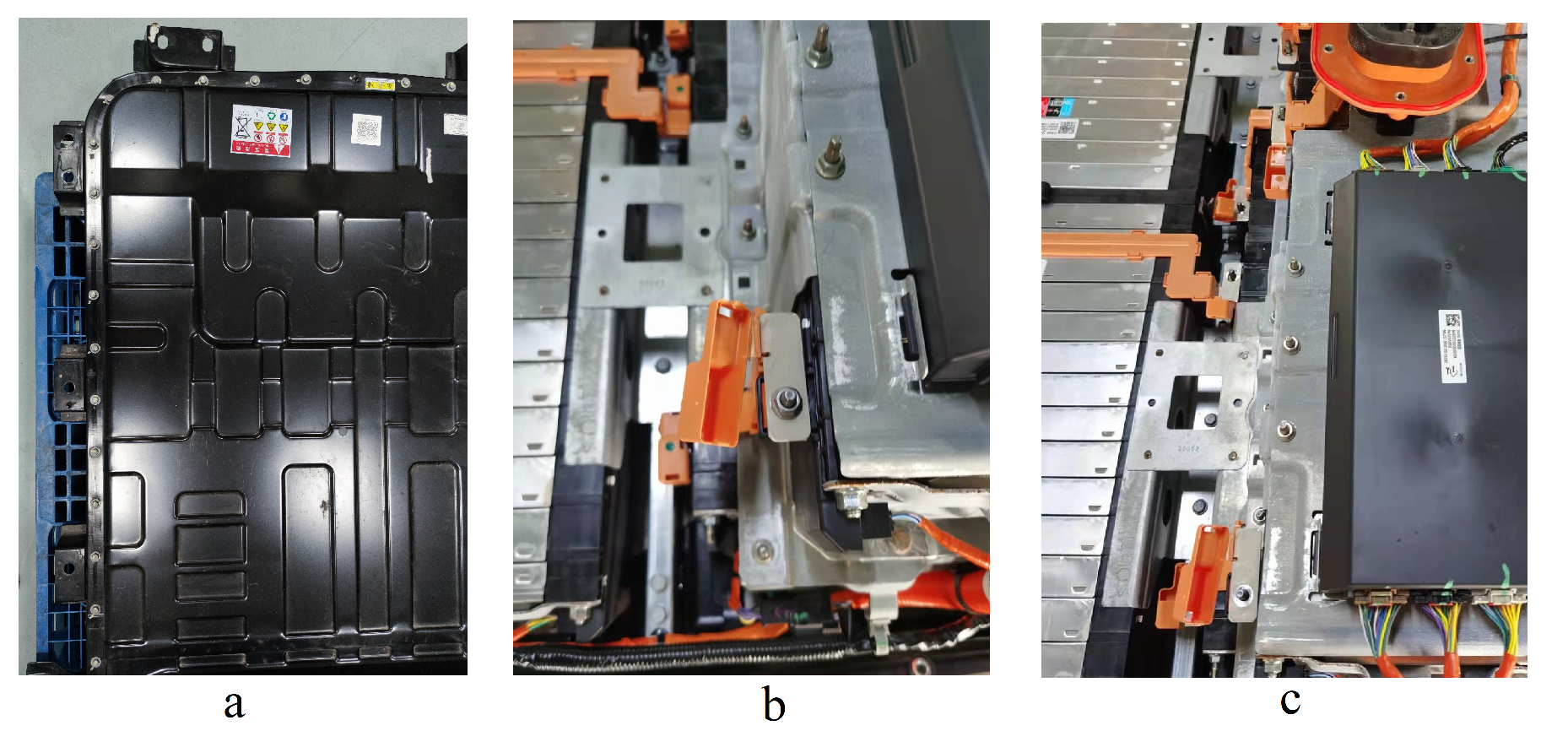}
\caption{Bolt Fastening Connections of Power Battery.}
\label{fig:bolt fastening}
\end{figure}

Disassembly is necessary for echelon utilization and resource regeneration of power batteries. In the manufacturing process of power battery, bolt fastening connection is widely used in many circumstances, such as fastening of power battery shells(Figure \ref{fig:bolt fastening}a), fastening of bus plates(Figure \ref{fig:bolt fastening}b), and fastening of battery modules(Figure \ref{fig:bolt fastening}c). Therefore, this paper focuses on bolt disassembly, the most common and typical task of power battery disassembly, as the initial investigation into battery disassembly tasks using NeuroSymbolic TAMP.
\subsection{Disassembly Tasks}
Disassembling bolts are straightforward to humans while challenging for robots to reliably and autonomously accomplish with the presence of environmental uncertainty. The goal is to design an algorithm to let robots autonomously carry out bolt disassembly tasks. To simulate environment dynamics, we use random obstacles around the bolts to test if the robot can choose the right action under different circumstances. This is critical to achieving the autonomous disassembly goal. 

Through analyzing the characteristics and focuses of the bolt disassembly process \cite{ref_13}, we designed five action primitives for the task: Approach, Mate, Push, Insert, Disassemble.

\textbf{(1)Approach}
This primitive directs the robot arms’ nut runner to approach the bolt from above. This primitive requires the bolt’s general location, which could be coarse, and only changes the robot arm’s position with no interactions with the target bolt.

\textbf{(2)Mate}
This primitive adjusts the robot arm’s nut runner to get its centerline aligned with the bolt’s centerline.

\textbf{(3)Push}
If perception indicates that there is not enough room for the nut runner to conduct disassembly, this primitive will be invoked to push away obstacles around the target. 

\textbf{(4)Insert}
This primitive directs the nut runner to rotate around the bolt axis to ensure valid contact, normally with a contact depth of 1mm to the gasket under the bolt. Meanwhile, the nut runner slowly rotates clockwise, ready to stop if a torque larger than 5Nm is detected to avoid anomalies.  

\textbf{(5)Disassemble}
This primitive directs the nut runner to rotate counterclockwise and move away from the bolt at a speed equivalent to the amount of pitch that the bolt moves per second. 

To achieve autonomous disassembly under dynamic and unstructured environments, the robot needs to choose appropriate primitives and sequences autonomously based on the current situation.
\subsection{Formulation of the disassembly task}
We formulate the bolt disassembly task as a planning issue ($S_0$,$S_G$,A), $S_0$ as the initial status of the disassembly system, $S_G$ as its target status, A as a sequence of disassembly primitives A = {a}, which consists of primitives from the aforementioned primitives set (Approach, Mate, Push, Insert, Disassemble).

Each primitive is defined by a set of descriptors: action(a), pre(a), eff(a), param(a).  Action(a) denotes the name of the primitive, with its required parameters described in param(a). Pre(a)defines this primitive's prerequisite conditions, while eff(a) represents this primitive's effects after being carried out.

Planning techniques focus on finding a disassembly primitive sequence (plan = ($a_1$,$a_2$,...,$a_i$,...,$a_n$)), and by executing these primitives in turn, the system moves from its initial status $S_0$ to the target status $S_G$. PDDL is a standard language to describe this formulation. This paper uses PDDL to define different disassembly status and actions, and based on PDDL implements autonomous battery disassembly.
\section{NeuroSymbolic TAMP}
As mentioned earlier, the battery disassembly issue faces the same challenges that NeuroSymbolic methods are trying to address. Classic planning methods use predicates to indicate the status. For example, the True/False value of predicate \emph{clear(A)} could be used to represent whether anything is 'above' object A and facilitate the following inference. This paper introduces neural predicates, using the neural network’s softmax output to indicate the system status, to address the classic planning problem. Neural predicate results are further output to the inference module and to define primitives. 
\subsection{Neural Predicates}
Predicate logic has been extensively adopted owing to its strong descriptive capabilities \cite{ref_14}. Predicates, as the core concept of predicate logic, are used to describe relationships between events and objects. The classical method refines the predicate definition continuously until the predicate corresponds to the sensor data to complete the relational representation. Defining predicates in this manner requires expert knowledge and is only applicable in restricted scenarios, which makes it unsuitable for dynamic and open environments.

To address these challenges this paper introduces a neural predicate, which is both a predicate and a neural network. It is responsible to convert continuous status to symbols that could be exploited by primitives. This method does not require complicated rules defined by experts beforehand. We use \emph{p(sensor,O)}to denote the neural predicates. Neural predicates take sensory data as the input. If a neural predicate has n groundings, then we use an n-dimensional vector as the output to represent this predicate's probabilistic distribution across n categories. For example, if a neural predicate $clear$ could be $True$ or $False$, then the output will be $[p_1,p_2]$ that corresponds to the probabilities of \emph{clear(sensor,True)} and \emph{clear(sensor,False)}, respectively. Deep learning's softmax output matches well with the n-dimensional probabilistic distribution, hence adopted as the output.

\begin{figure}[h]
\centering
\includegraphics[width=12cm]{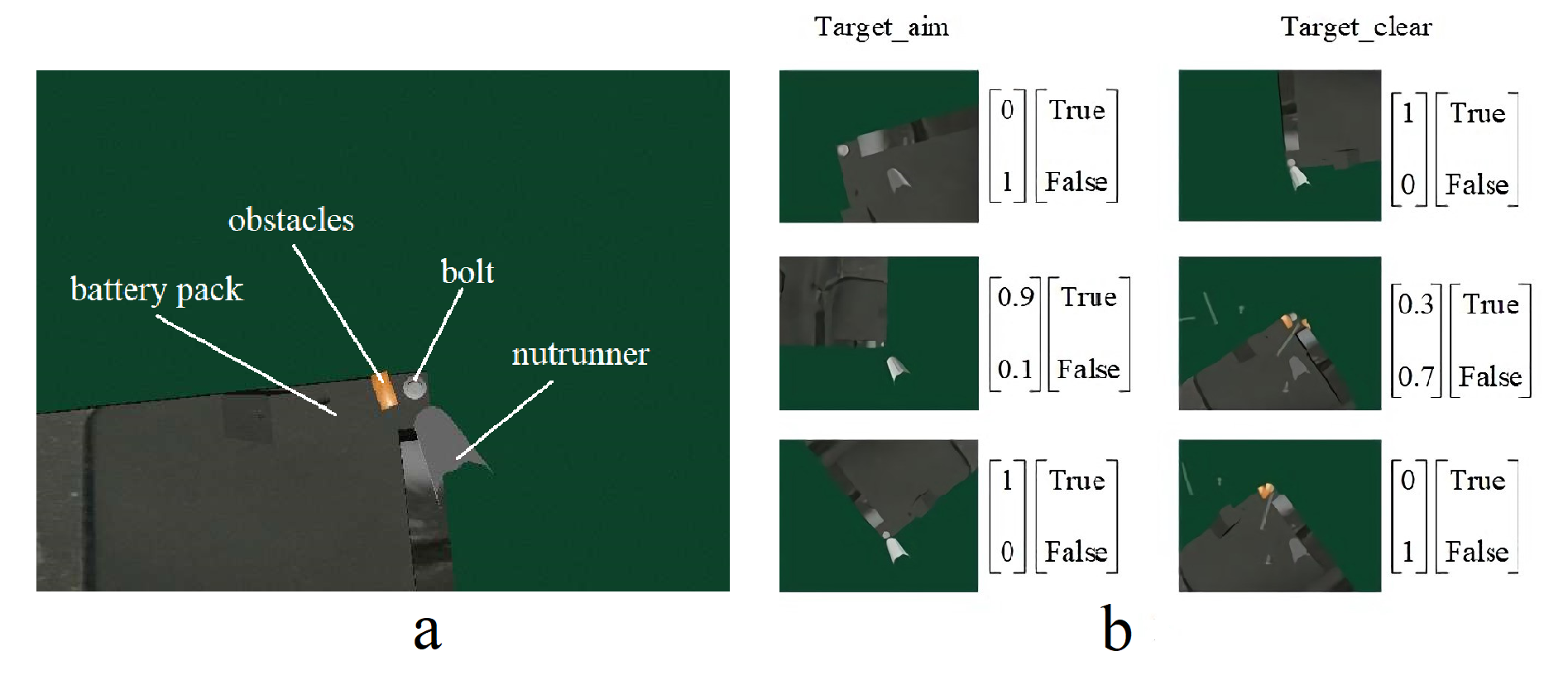}
\caption{Neuro Predicates Definition.}
\label{fig:two}
\end{figure}

We define two neural predicates for the bolt disassembly system (shown in ''Figure \ref{fig:two}''): ‘$target\_aim$’ to indicate target aligned and ‘$target\_clear$’ to indicate that there are no obstacles. 

It is worth mentioning that since neural predicates classifications are defined for a specific scenario, neural network training becomes much easier. In this paper's tests, we use 400 pictures to train each neural predicate and start with the pre-trained VGG-16 \cite{ref_15} for the training. Predicate ‘$target\_aim$’ achieves an accuracy of 98\% after 40 epochs of training, while predicate '$target\_clear$' achieves an accuracy of 96\% after 40 epochs of training. Such accuracy well suffices to serve the following inferring tasks.

\subsection{Disassembly Primitives}
In robot task planning scenarios, it's the primitives that bridge logic planning and the robot's physical movements. PDDL can accurately describe in symbolic space each primitive and its prerequisite (:pre), effects after execution (:eff). It also unambiguously describes the system's initial and goal states. In the symbolic space, the PDDL-based planner infers the primitive sequence from the initial state (Init) to the target state (Goal) as the action plan. Battery disassembly, specifically bolt disassembly tasks, involves primitives including Approach, Push, Mate, Insert, and Disassemble, etc. Their corresponding PDDL definitions are as follows:
\begin{lstlisting}[language=PDDL]
(:Init have(coarse_pose)
(:Goal disassembled(sensor)

(:action Approach
:param (coarse_pose sensor)
:pre(have(coarse_pose)
:eff(and(above_bolt)(target_aim(sensor))
(target_clear(sensor)))

(:action Mate
:param (sensor)
:pre(and(above_bolt(sensor))(not(target_aim(sensor)))
:eff(target_aim(sensor)))

(:action Push
:param (sensor)
:pre(and(above_bolt(sensor))
(not(target_clear(sensor))))
:eff (target_clear(sensor))

(:action Insert
:param (sensor)
:pre(and(target_aim(sensor))(target_clear(sensor)))
:eff(cramped(sensor))

(:action Disassemble
:param (sensor)
:pre(cramped(sensor))
:eff (disassembled(sensor))
\end{lstlisting}

The system defines for each primitive its controller. Primitive controllers control robots to execute in the physical environment which is prescribed by the primitives, i.e. when prerequisite (:pre) is satisfied, using parameters (:param) as input, to control the robot to transit from its current status to status described in (:eff). For the sake of testing, this paper uses ROS Moveit to develop primitive controllers. Note that primitive controllers can be implemented by many other ways. 

\subsection{Disassembly Planning}

With abstractions supported by neural predicates and disassembly primitives, the original complicated planning problem in the continuous space is converted to finding the best action plan in the symbolic space. By following the action plan sequentially, the disassembly system makes the robot transfer from its initial status $S_0$ to its target status $S_G$. We use forward searching to address this issue. Algorithm 1 shows the specific algorithm.

Even though the neural predicates and probabilistic representations of status are introduced to avoid inconsistencies caused by manually defined thresholds, there still exist planning uncertainties. For example, after approaching the target object by executing initial locating primitives, is it still necessary to adjust robot poses? While executing the “Push” primitive, is it ensured that obstacles are fully cleared up? To minimize human interventions, after the execution of each primitive, the system will ensure that it has achieved the desired results. If not, the system will re-execute the planning algorithm to obtain a new action plan based on the current status and the goals.

\SetKwComment{Comment}{/* }{ */}
\begin{algorithm}
\scriptsize
\caption{NeuroSymbolic TAMP for Disassembly Task }\label{alg:one}
\textbf{Input}:\ $S_0,S_G,A$
\Comment*[r]{Initial State, Target State, Primitive set}
\textbf{Output}:\ $op\_list$
\Comment*[r]{primitive sequence to accomplish the task}
\BlankLine
$initial\_queue(Q_1)$\;
$Q_1.enqueue((S_0,\{\} ))$\
\Comment*[r]{Initialize Q1, The initial state of the planning is $S_0$; \{\} is the list of disassemble action primitives, initial state is empty }
\While{$Q_1 \neq null$}
{$curr\_status,curr\_op\_list = Q_1.pop\_front()$\;
$initial\_set(tmp\_set)$ 
\Comment*[r]{Temporarily record the current state and the new state after using the operation primitive}
\For(\Comment*[r]{Traverse all of the action primitives definition in the planning problem}){$a \ in \ A$}
{\If{$curr\_status \supset a.precondition$}{
$new\_status=apply(curr\_status,a)$\Comment*[r]{Apply action primitives a to the current state}
$new\_op\_list=curr\_op\_list+a$\;
\If(\Comment*[f]{Achieve the goal state}){$new\_status \supset SG$}
{\Return $new\_op\_list$}
$tmp\_set.add((new\_status,new\_op\_list))$\;}}
$List \ L= sort\_and\_filter(tmp\_set)$\Comment*[r]{Sort the tmp set tuples by likelihood and filter out tuples with a low probability}
\For{$t\ in \ L$}
{$Q_1.enqueue(t)$\Comment*[r]{Add the tuples into the queue that meet the requirements }}
} 
\end{algorithm}

\section{Test Results}
To test the effectiveness of this NeuroSymbolic TAMP for the battery disassembly task, we conducted battery bolts disassembly tests both with and without obstacles on the robot simulation platform. Test results have shown that the methods proposed in this paper are effective indeed.
\subsection{Test Configurations}
\begin{figure}[ht]
\centering
\includegraphics[width=12cm]{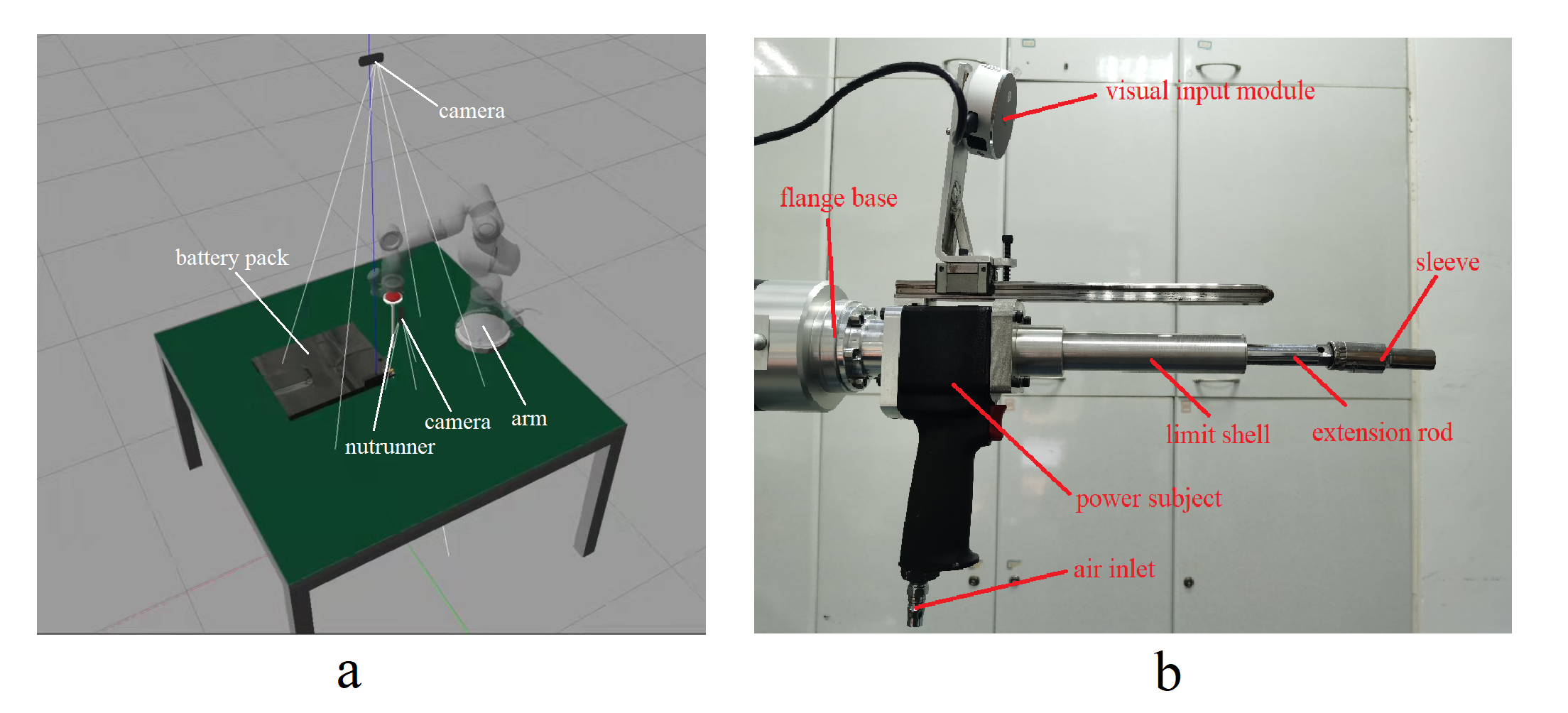}
\caption{Disassembly Platform. a: disassembly simulation environment; b: the nutrunner corresponding to the simulation enviroment}
\label{fig:three}
\end{figure}

Figure \ref{fig:three}a shows the test environment. We use a visual simulation environment Gazebo with a 7-DOF robot arm xMatePro7 ,a product of ROKAE, as the testing platform. 
The robot arm is equipped with a nutrunner at its end, which could be controlled by ROS MoveIt!. The nutrunner is actually a pneumatic torque actuator(as shown in Figure \ref{fig:three}b) which is composed of visual input module, power subject, sleeve, flange base and limit shell, etc \cite{ref_16}. In more detail, the visual module uses Intel RealSense LiDAR Camera L515 to estimate the position and pose of the target object, the power subject similar to the internal structure of the pneumatic wrench converts the kinetic energy of the high-pressure air into the kinetic energy of the sleeve rotation, the sleeve is mounted on the extension rod (the sleeve can be replaced in different sizes), the flange base of the actuator is fixed at the end of robot arm, high-pressure air enters the power subject through the air inlet, the limit shell aims to prevent the extension rod from swinging in a wide range. The overall structure is mainly used to realize the passive compliant disassembly of the bolt due to insufficient accuracy in the position and pose estimation of the bolt, compensating for the error by using mechanical structure.

In addition, the simulation platform is also equipped with a global camera (on top, overseeing the whole environment) and a regional camera (at the end of the robot arm). Camera parameters are consistent with Intel RealSense Depth Camera D435i. The simulator is also configured with a Chang’an motors EADO EV460 battery pack, bots, and a woodblock as the obstacle. All tests are run on an Intel NUC.

First, we observe if the proposed method could autonomously adapt to different scenarios with different task plans.  Next, we use SR (Success Rate) to evaluate the algorithm’s effectiveness against perception errors. The baseline is a classic method that treats perception and control as separate tasks and the control module assumes accurate enough target positions from perception. Note that the accurate target positions could be from either symbolic or neural methods. Here we intend to compare how perception errors affect the proposed and classical solutions. 
\subsection{Autonomous Task Planning}
\begin{figure}
\centering
\includegraphics[width=12cm]{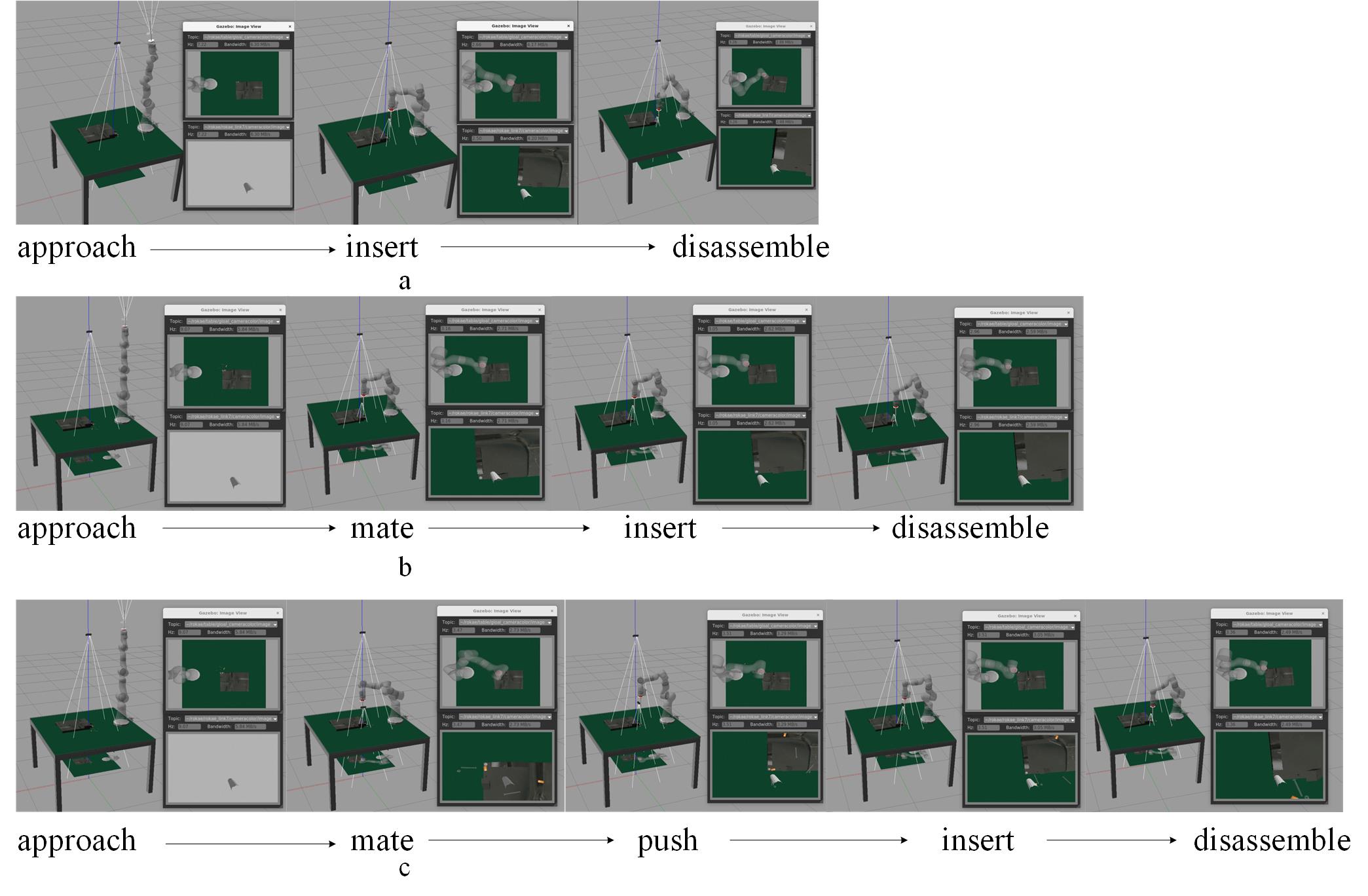}
\caption{Autonomous planning under different situations.}
\label{fig:four}
\end{figure}

Figure \ref{fig:four} shows three different situations that the robot might face during the task:

\begin{enumerate}
  \item The robot arm has accurate location information of the battery bolt; 
  \item The robot arm does not have accurate location information of the battery bolt;
  \item There are obstacles near the bolts.
\end{enumerate}

From Figure \ref{fig:four}, it can be seen that the proposed method autonomously adjusts the task plan to adapt to different situations. For case 1 it takes only 3 steps to accomplish the task. For case 2 since the robot does not have an accurate bolt location, it takes an extra step to make sure the nut runner and the bolt are lined up with each other properly. For case 3 due to both inaccurate location information and obstacles around the bolt, the system takes two extra steps to first line up the nut runner with the bolt and then clear the area around the bolt. The test result shows that the proposed method can autonomously plan and execute based on the perceived situation.
\subsection{Tests Results without Obstacles}
To test the robustness of NeuroSymbolic TAMP against inaccurate target locations for battery disassembly task, we randomly position the target bolt on the platform and then introduce an error with $N(0,\sigma ^{2})$ distribution to the position. The tests are repeated with different $\sigma$ to obtain SR.

\begin{figure}[h]
\centering
\includegraphics[width=12cm]{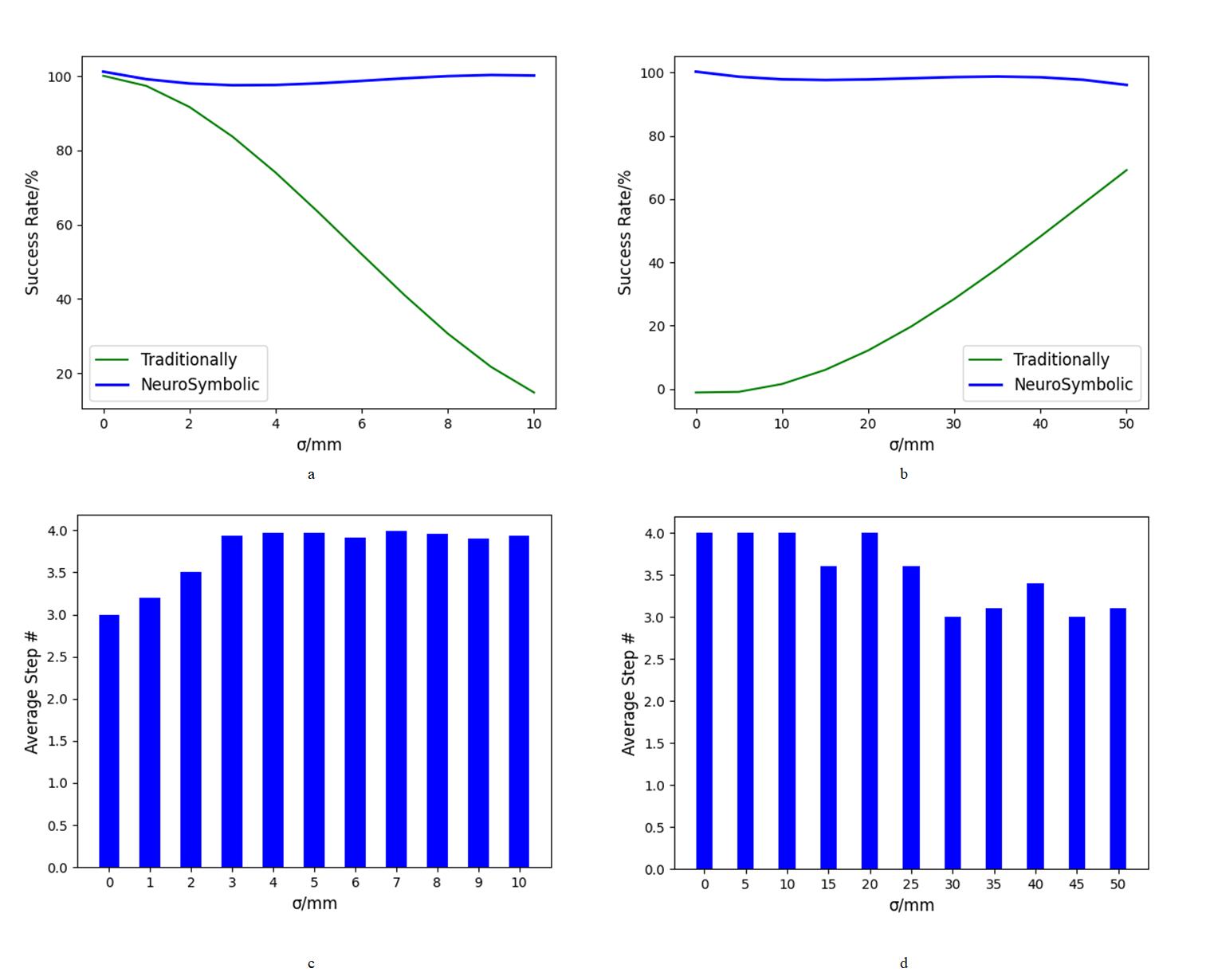}
\caption{Test data. a: success rate of obstacle-free disassembly tests; b: the success rate of disassembly test with obstacles; c: average steps number of obstacle-free disassembly tests; d: average number of steps for disassembly test with obstacles.}
\label{fig:six}
\end{figure}

Figure \ref{fig:six}a shows the test results: The classic method \cite{ref_22} achieves a decent SR while the variance is less than 1mm, which its performance quickly deteriorates as the variance increases. The baseline method achieves decent performance in a static and structured environment, which reflects the current industry status. Unstructured environments pose a completely new set of challenges to the classic methods, though. In comparison, the proposed method can adapt and adjust based on online scenarios to achieve a nearly 100\% SR. Among the 400 trials, the 3 Failures are due to inadequate error handling of certain primitives (deadlock for the system cannot find the bolt's 6 corners under poor lighting conditions). The high success rate of the proposed method is mainly due to its ability to execute the $Mate$ action autonomously (as shown in Figure \ref{fig:six}c). The experimental results show the broad prospect of man-machine cooperation due to high success rate and enhanced security compared to the existing solution \cite{ref_4}.

\subsection{Test Results with Obstacles}
This is to test the proposed method's effectiveness with obstacles. The tests are conducted randomly initializing the target bolt's position on the platform, then use this location with a deviation of $N(0,\sigma^{2})$ distribution as the obstacle's location. Obstacles include bolts, nuts, and wooden blocks. The tests are conducted with different $\sigma$ to calculate the SR.

Figure \ref{fig:six}b shows the test results. The classic method is not capable of cleaning up obstacles, hence its poor SR when the obstacles are of smaller $\sigma$ (the obstacles are highly likely to block the bolts) and better SR with large $\sigma$. This is exactly what existing solutions need to be improved and what the proposed method intends to address. The method proposed by this paper achieves a high SR consistently, regardless of the $\sigma$ value. It should be emphasized that the proposed method does not execute '$Push$' every time, as shown in Figure \ref{fig:six}c. It autonomously decides whether '$Push$' is necessary based on the current status. When $\sigma$ is small, i.e., it is more likely that the obstacles are in the way, the system will execute this action sequence: $Approach \to Push \to Insert \to Disassemble$. When $\sigma$ is large, i.e., it is less likely that the obstacles are in the way, the system will execute this sequence of 3 actions: $Approach \to Insert \to Disassemble$. From this perspective, we can highlight the flexibility of the proposed method with better adaptation to the real site compared to the pre-programming solution \cite{ref_17}.
\section{Summary and Future Work}
Due to high uncertainties, it is challenging to pre-plan for battery disassembly tasks in a complex and unstructured working environment. This paper proposes NeuroSymbolic TAMP, an autonomous planning method based on NeuroSymbolic computing, using the collaborative robot equipped with an adapted pneumatic nutrunner. Disassembly primitives are introduced to bridge logic planning and the robot's physical movements, which describe the system's initial and goal states. The definition of neural predicates is responsible to convert continuous status to symbols that could be exploited by disassembly primitives. Finally, Tests were conducted on a visual simulation environment Gazebo to validate and demonstrate the feasibility of the proposed methodology, which have shown that it works autonomously and effectively with an SR of $98\%$. 

But this is only the first step in building an intelligent system for battery disassembly. The developed method has several issues requiring attention: (1) transformation from simulation to real machine for testing needs to be conducted to lay a solid foundation for the practical application of this technology with improved reliability and safety; (2)
we need to extend the action primitives and predicates so that the robot can complete more complex tasks such as upper cover removal and bus bar removal; (3) as robots can achieve more and more jobs, knowledge graphs should be introduced to manage all kinds of information required by NeuroSymbolic TAMP making full use of accumulated domain knowledge and accomplish goals intelligently; (4) challenges regarding the deployment of NeuroSymbolic TAMP need more attention.

Future work will address the above issues by optimizing relevant algorithms and perception ability to transform from simulation to real machine. The action primitives and predicates will be extended to complete more tasks in the real battery disassembly process. More experiments will be carried out, and possible failure models of the designed method will be investigated to enhance its robustness.

%
%
%


\begin{thebibliography}{99}
%

\bibitem{ref_1}
Dixon, J., Bell, K.:Electric vehicles: Battery capacity, charger power, access to charging and the impacts on distribution networks. eTransportation. vol. 4, pp. 100059(2020). \url{doi:10.1016/j.etran.2020.100059}

\bibitem{ref_2}
Harper, G., Sommerville, R., Kendrick, E., Driscoll, L., Slater, P., Stolkin, R., Walton, A., Christensen, P., Heidrich, O., Lambert, S., Abbott, A., Ryder, K., Gaines, L., Anderson, P.:Recycling lithium-ion batteries from electric vehicles. Nature. vol. 575(7781), pp. 75-86(2019). \url{doi:10.1038/s41586-019-1682-5}

\bibitem{ref_3}
Rastegarpanah, A., Gonzalez, H.C., Stolkin, R.:Semi-Autonomous Behaviour Tree-Based Framework for Sorting Electric Vehicle Batteries Components. Robotics. vol. 10(2), pp. 82(2021). \url{doi:10.3390/robotics10020082}

\bibitem{ref_4}
Choux, M., Marti Bigorra, E., Tyapin, I.:Task Planner for Robotic Disassembly of Electric Vehicle Battery Pack. Metals. vol. 11(3), pp. 387(2021). \url{doi:10.3390/met11030387}

\bibitem{ref_17}
Wegener, K., Andrew, S., Raatz, A., Dröder, K., Herrmann, C.:Disassembly of Electric Vehicle Batteries Using the Example of the Audi Q5 Hybrid System. Procedia CIRP. vol. 23, pp. 155-160(2014). \url{doi:10.1016/j.procir.2014.10.098}

\bibitem{ref_18}
Zude, Z., Jiayi, L., Truong Pham, D., Wenjun, X.,Javier Ramirez, F., Chunqian, J., Quan, L.:Disassembly sequence planning: Recent developments and future trends. Proceedings of the Institution of Mechanical Engineers, Part B: Journal of Engineering Manufacture. vol. 233, pp. 1450-1471(2018). 
\bibitem{ref_19}
Ong, S. K., Chang, M. M. L., Nee, A. Y. C.:Product disassembly sequence planning: state-of-the-art, challenges, opportunities and future directions. International Journal of Production Research. vol. 59(11), pp. 3493-3508(2021). \url{doi:10.1080/00207543.2020.1868598}

\bibitem{ref_20}
Hellmuth, Jan F., DiFilippo, Nicholas M., Jouaneh, Musa K.:Assessment of the automation potential of electric vehicle battery disassembly. Journal of Manufacturing Systems. vol. 59, pp. 398-412(2021). \url{doi:10.1016/j.jmsy.2021.03.009}

\bibitem{ref_21}
Liurui, L., Panni, Z. Tairan, Y., Sturges, R., Ellis, Michael W., Zheng, L.:Disassembly Automation for Recycling End-of-Life Lithium-Ion Pouch Cells. JOM. vol. 71(12), pp. 4457-4464(2019). \url{doi:10.1007/s11837-019-03778-0}

\bibitem{ref_22}
Garrett, C. R., Chitnis, R., Holladay, R., Kim, B., Silver, T., Kaelbling, L. P., Lozano-Pérez, T.:Integrated Task and Motion Planning. Annual Review of Control, Robotics, and Autonomous Systems. vol. 4(1), pp. 265-293(2021). \url{doi:10.1146/annurev-control-091420-084139}

\bibitem{ref_23}
Castaman, N., Pagello, E., Menegatti, E., Pretto, A.:Receding Horizon Task and Motion Planning in Changing Environments. arXiv:2009.03139. 2021.

\bibitem{ref_24}
Haslum, P., Lipovetzky, N., Magazzeni, D., Muise, C.:An Introduction to the Planning Domain Definition Language. Morgan \& Claypool. 2019.

\bibitem{ref_25}
Serrano, S. A., Santiago, E., Martinez-Carranza, J., Morales, E. F., Sucar, L. E.:Knowledge-Based Hierarchical POMDPs for Task Planning. Journal of Intelligent \& Robotic Systems. vol. 101(4), pp. 82(2021). \url{doi:10.1007/s10846-021-01348-8}

\bibitem{ref_26}
Dicong Q., Yibiao Z., Chris L. B.:Latent Belief Space Motion Planning under Cost, Dynamics, and Intent Uncertainty. In: Proceedings of Robotics: Science and Systems. Corvalis, Oregon, USA(2020).

\bibitem{ref_27}
Sarantopoulos, I., Kiatos, M., Doulgeri, Z., Malassiotis, S.:Total Singulation With Modular Reinforcement Learning. IEEE Robotics and Automation Letters. vol. 6(2), pp. 4117-4124(2021). \url{doi:10.1109/LRA.2021.3062295}

\bibitem{ref_28}
Garrett, C. R., Paxton, C., Lozano-Pérez, T., Kaelbling, L. P. Fox, D.:Online Replanning in Belief Space for Partially Observable Task and Motion Problems. In: 2020 IEEE International Conference on Robotics and Automation (ICRA). Paris, France(2020).

\bibitem{ref_5}
H{\aa}kan L. S. Younes, Michael L. Littman.:PPDDL 1.0 : An Extension to PDDL for Expressing Planning Domains with Probabilistic Effects. Technical Report, 2004.

\bibitem{ref_6}
Fierens, D., Broeck, G., Renkens, J., Shterionov, D., Gutmann, B., Thon, I., Janssens, G., De Raedt, L.:Inference and learning in probabilistic logic programs using weighted Boolean formulas. Theory and Practice of Logic Programming. vol. 15, pp. 358-401(2014). \url{doi:10.1017/s1471068414000076}

\bibitem{ref_7}
De Raedt, L., Kimmig, A., Toivonen, H.:ProbLog: A Probabilistic Prolog and Its Application in Link Discovery. In: Proceedings of 20th International Joint Conference on Artifical Intelligence, pp. 2468–2473. Morgan Kaufmann, Hyderabad, India(2007).

\bibitem{ref_8}
Artur d'Avila Garcez, Luis C.Lamb.:Neurosymbolic AI: The 3rd Wave. arXiv:\\
2012.05876. 2020.

\bibitem{ref_9}
Daniel K.: Thinking, Fast and Slow. Farrar, Straus and Giroux, 2013.

\bibitem{ref_10}
Sarantopoulos, I., Kiatos, M., Doulgeri, Z., Malassiotis, S.:Total Singulation With Modular Reinforcement Learning. IEEE Robotics and Automation Letters. vol. 6(2), pp. 4117-4124(2021). \url{doi:10.1109/LRA.2021.3062295}

\bibitem{ref_11}
Jiayuan M., Chuang G., Pushmeet K., Joshua B.T., Jiajun W.:The Neuro-Symbolic Concept Learner: Interpreting Scenes, Words, and Sentences From Natural Supervision. arXiv:1904.12584. 2019

\bibitem{ref_12}
De Raedt, L., Manhaeve, R., Dumancic, S., Demeester, T., Kimmig, A.:Neuro-Symbolic = Neural + Logical + Probabilistic. In: Proceedings of the 2019 International Workshop on Neural- Symbolic Learning and Reasoning. Macao, China(2019).

\bibitem{ref_13}
Li, R., Pham, D.T., Huang, J., Tan, Y., Qu, M., Wang, Y., Kerin, M., Jiang, K., Su, S., Ji, C., Liu, Q., Zhou, Z.:Unfastening of Hexagonal Headed Screws by a Collaborative Robot. IEEE Transactions on Automation Science and Engineering. vol. 17(3), pp. 1455-1468(2020). \url{doi:10.1109/TASE.2019.2958712}

\bibitem{ref_14}
Van Emden, M.H., Kowalski, R.A.:The Semantics of Predicate Logic as a Programming Language. J. ACM. vol. 23(4), pp. 733–742(1976). \url{doi:10.1145/321978.321991}

\bibitem{ref_15}
Simonyan, K., Zisserman, A.:Very Deep Convolutional Networks for Large-Scale Image Recognition. arXiv:1409.1556. 2015.

\bibitem{ref_16}
Shanghai Yongxu Technology Co., Ltd. Passive compliant pneumatic torque actuator at the end of robot, China Patent CN202110783210.2.

\end{thebibliography}
\end{document}